# Intelligent Co-Design: An Interactive LLM Framework for Interior Spatial Design via Multi-Modal Agents

Ren Jian Lim[1], Rushi Dai[1*]

[1]Hong Kong Center for Construction Robotics, The Hong Kong University of Science and Technology

**Abstract**

In architectural interior design, miscommunication frequently arises as clients lack design knowledge, while designers struggle to explain complex spatial relationships, leading to delayed timelines and financial losses. Recent advancements in generative layout tools narrow the gap by automating 3D visualizations. However, prevailing methodologies exhibit limitations: rule-based systems implement hard-coded spatial constraints that restrict participatory engagement, while data-driven models rely on extensive training datasets. Recent large language models (LLMs) bridge this gap by enabling intuitive reasoning about spatial relationships through natural language. This research presents an LLM-based, multimodal, multi-agent framework that dynamically converts natural language descriptions and imagery into 3D designs. Specialized agents (Reference, Spatial, Interactive, Grader), operating via prompt guidelines, collaboratively address core challenges: the agent system enables real-time user interaction for iterative spatial refinement, while Retrieval-Augmented Generation (RAG) reduces data dependency without requiring task-specific model training. This framework accurately interprets spatial intent and generates optimized 3D indoor design, improving productivity, and encouraging nondesigner participation. Evaluations across diverse floor plans and user questionnaires demonstrate effectiveness. An independent LLM evaluator consistently rated participatory layouts higher in user intent alignment, aesthetic coherence, functionality, and circulation. Questionnaire results indicated 77% satisfaction and a clear preference over traditional design software. These findings suggest the framework enhances user-centric communication and fosters more inclusive, effective, and resilient design processes. Project page: https://rsigktyper.github.io/AICodesign/

## 1. Introduction

In recent years, the need for automation in architectural design has become widely recognized. Automation streamlines complex, time-consuming workflows, letting computers handle repetitive tasks so designers can concentrate on creativity and concept development (Shi et al. 2024; Jiang,

---

Wang, and Ma 2023; Matter and Gado 2024). Advances in computer science have produced powerful techniques for synthetic 3D scene generation, yet most overlook participatory design, even as they shorten scene creation into minutes and boost efficiency.

Mainstream scene synthesis methodologies remain indifferent toward participatory engagement. Rule-based systems such as Infinigen Indoors (Raistrick et al. 2024) generate realistic 3D scenes but embed hard-coded spatial constraints that resist detailed client customization. Data-driven frameworks similar to SceneDreamer (Chen, Wang, and Liu 2023) replace fixed rules with learning-based models, increasing flexibility at the cost of extensive training datasets. LLM-based tools such as Holodeck (Yang et al. 2024) reduce data demands and broaden layout diversity but still presume command-line knowledge from users. Efficiency improvements, therefore, often come at the expense of inclusivity.

Large language models (LLMs) combined with agentic systems offer a remedy. Recent LLM agents combine natural language dialogue with retrieval-augmented knowledge (RAG) and few-shot adaptability, enabling stakeholders to take advantage of pretrained models without heavy data overhead. By decomposing user requirements into specialized subtasks, agent systems can translate nondesigner intent into computable spatial rules while maintaining conversational flow. Building on these insights, we propose an LLM-based, multimodal, multi-agent system that strengthens stakeholder interaction and increases productivity. Guided by Norman's (2013) fundamental principles, namely affordances, signifiers, mappings, feedback, and coherent conceptual models, we design our multi-agent framework to ensure user control and rapid feedback, ultimately enhancing participatory design.

The framework arranges four agents together. The Reference Agent derives spatial relationships from images; the Spatial Agent proposes furniture categories and spatial rules; the Interactive Agent interprets those rules into plain language for real-time user interaction; and the Grader Agent scores generated spatial arrangements for refinement. The optimized spatial rules are exported as a 3D scene, accompanied by supplementary log files. Our research then evaluates the framework with an independent LLM that scores user intent alignment, aesthetic coherence, functionality, and circulation, and with a user questionnaire that measure satisfaction and tool preference.

In summary, our contributions are as follows: (1) we present a robust AI-driven framework for indoor design built on multimodal, multi-agent systems; (2) we demonstrate resiliency of the framework by generating layouts across diverse floor plan typologies without additional training data; (3) we present a detailed real-world application showing how the framework converts nondesigner intent into 3D scenes; (4) we discuss pathways for extending the approach to broader architectural scenarios, paving the way for more inclusive and efficient design workflows.

## 2. Related Works

### 2.1 AI in Architectural Interior Design

Artificial intelligence is reshaping architectural design by improving generative workflows, optimizing spatial layouts, and bridging communication between designers and clients. DiffDesign (Yang et al. 2024) demonstrates controllable diffusion that converts text or sketches into high-quality room images but remains confined to static 2D output and depends on large training sets.

"Synthesizing User Preferences from Supplier Catalogs" (Kuang et al. 2024) advances to "2D plus" visualization by integrating CLIP retrieval, Grounding DINO, and Stable Diffusion, enabling real-time edits such as changing sofa fabric to match commercial catalogs, but still limits feedback to images. Spacify (Vaidhyanathan, Radhakrishnan, and Castillo y López 2023) shifts focus from imagery to 3D space, combining GPT-4 with augmented reality so users can walk through generated layouts; however, a preliminary baseline room scan is required. Collectively, these studies prove AI can accelerate interior workflows and broaden participation, yet each approach faces its own trade-offs.

### 2.2 Automation in 3D Scene Generation

Research about scene generation in computer graphics converts text, point clouds, or graphs into full 3D environments. Text2NeRF (Zhang et al. 2024) turns natural-language prompts into Neural Radiance Fields; "3D Scene Graph Generation from Point Clouds" (Wei et al. 2024) extracts relational graphs directly from point clouds; and GraphDreamer (Gao et al. 2024) creates compositional scenes from symbolic graphs. While flexible, all three depend on sizable training datasets. Infinigen Indoor (Raistrick et al. 2024) eliminates such bottlenecks through procedural rules and constraint graphs, producing photorealistic interiors without external training. Holodeck (Yang et al. 2024) goes a step further, integrating large language models with embodied agents so users can reshape scenes through natural language. However, domain-specific challenges persist. Hard-coded constraints in Infinigen hinder personalized edits, whereas Holodeck demands command-line knowledge, reducing accessibility for nontechnical stakeholders. An ideal system would blend rule-based reliability with natural-language flexibility while avoiding heavy data requirements.

### 2.3 Co-design and Agent Systems

In participatory design, co-design positions end users as "experts of their experience" and relies on generative tools to support ideation and expression (Sanders and Stappers 2008; Sleeswijk Visser et al. 2005). In parallel, Schön (1983) describes "reflection-in-action" as practitioners, often prompted by surprise, turning thought back on action and on tacit knowing-in-action, surfacing and restructuring understandings as they proceed. These perspectives inform interactive systems that emphasize iterative refinement alongside technical capability. Building on these principles, recent AI agent systems translate user intent into actions and provide immediate feedback.

LLM-based agents are interactive systems capable of perceiving their environment, processing language inputs, and acting autonomously (Franklin and Graesser 1997; Wang et al. 2024). Early work such as ELIZA demonstrated conversational interaction but relied on rule-based patterns (Weizenbaum 1966; Caldarini, Jaf, and McGarry 2022). Recent large language models enable more diverse interaction, yet complex tasks often benefit from specialization. "Chain-of-Experts" (Xiao et al. 2024) shows that dividing work among specialized LLM agents outperforms a single agent. In interior design, I-Design (Çelen et al. 2025) exemplifies this approach by assigning distinct agents to select objects, propose spatial relations, and assemble scene graphs, simplifying iterative refinements. Designing such systems requires care: common failure examples include vague role definitions and weak inter-agent communication. "Why Do Multi-Agent LLM Systems Fail" (Pan et al. 2025) catalogs these pitfalls and suggest tactical and structural safeguards, offering guidance for more robust multi-agent pipelines.

## 3. Methods

This section introduces a multimodal, multi-agent framework for indoor spatial design generation (Fig. 1). Built on Claude 4.0 Sonnet (Anthropic n.d.) and the vision-language model LLaVa (Liu et al. 2024), the framework integrates conversational text and reference imagery to enable real-time participatory design. The framework positions users as "experts of their experience," and the agents expose editable spatial relationships with immediate feedback to support reflection-in-action (Sanders and Stappers 2008; Schön 1983). Overall, four specialized agents operate collaboratively: the Reference Agent extracts spatial information from images; the Spatial Agent selects furniture categories and generates corresponding spatial rules; the Interactive Agent converts spatial rules to natural language; and the Grader Agent scores the spatial rules. The refined layout is exported as a 3D scene, accompanied by log files and generation metrics. Claude was selected for its code-oriented reasoning and broad regional availability, while LLaVa offers cost-free, fine-grained image interpretation. By integrating LLM agents and automating repetitive tasks, designers are allowed to focus on creative strategy and delivering robust layouts matching varied client demands. Task distribution across specialized agents keeps stakeholders engaged, enhances communication, and shortens project turnaround. In the following subsections, we present details of each agent, the prompt engineering techniques that secure consistent LLM output, and integration of our framework into an existing 3D scene generator, demonstrating the application of our framework to indoor spatial design.

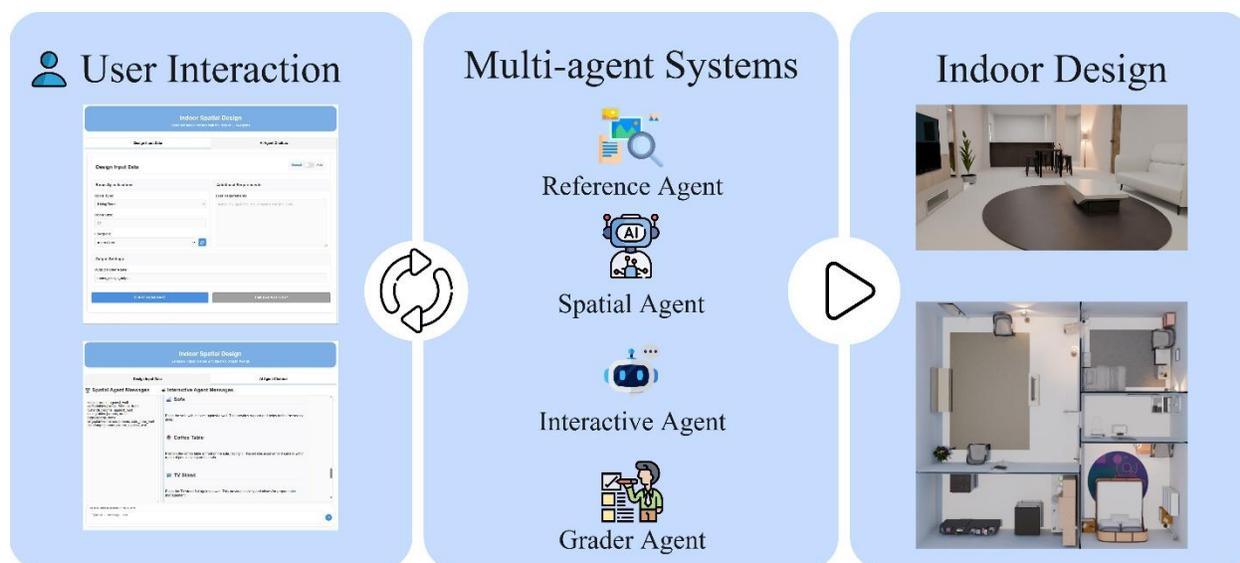

**Figure 1. Framework overview: a frontend interface collects input from user-agent dialogue; these inputs flow to a multi-agent system that iteratively refines spatial rules and output the final 3D interior scene.**

## 3.1 Interaction Between System and Users

Our framework emphasizes participatory design by providing users three decision stages (object selection, object constraints, and object score terms) during layout generation. As shown in Fig. 2, at each stage the Interactive Agent paraphrases the raw output of the Spatial Agent into plain language on the user interface. Users review, modify, accept, or reject. Acceptance locks the decision and passes it downstream; rejection returns feedback, prompting the Spatial Agent to regenerate until approval. Participants who prefer a hands-off route can toggle "auto mode," delegating review to the Grader Agent, which scores proposals against reference images. This loop keeps users in control without demanding technical fluency while still providing a fully automated fallback. Continuous feedback ensures that user intent remains central, the Spatial Agent adapts to evolving preferences, and overall design time is minimized. The resulting workflow combines inclusivity with efficiency, strengthening both communication and outcome quality in indoor spatial projects.

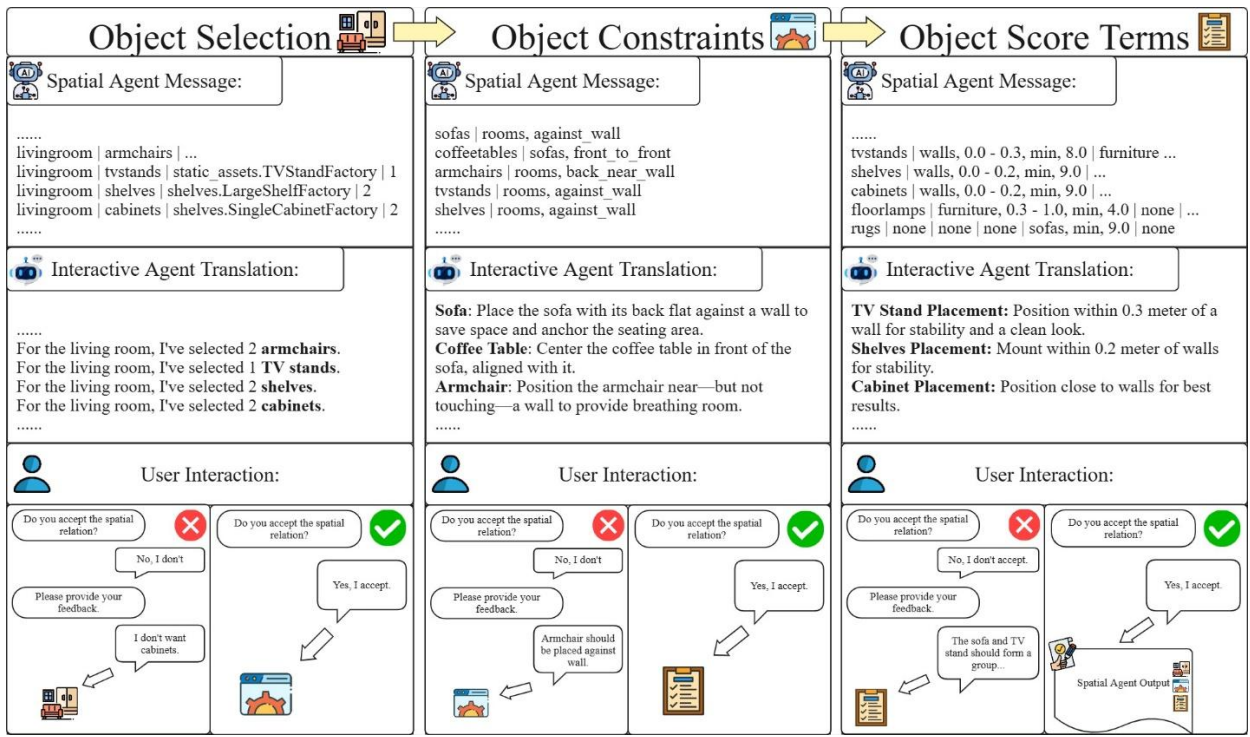

**Figure 2. Diagram of user-agent interaction across three stages: object selection, constraints, and score terms.**

## 3.2 Interaction Between Agents

The interactions between the agents within our framework are designed carefully to maximize performance (Fig. 3). The information flow begins with the Reference Agent, which provides foundational descriptions guiding the Spatial Agent's initial design decisions. The Spatial Agent, empowered by user input and RAG, generates structured outputs for furniture object selection and object spatial relation. Here, interaction between agents will divide into two modes: manual mode and auto mode. In manual mode, the Interactive Agent will translate raw outputs generated from the Spatial Agent into human-friendly text description, then display these messages to the front-end interface. In auto mode, the Grader Agent will evaluate the raw spatial output based on some good design images as grading reference. Both modes will optimize the spatial rule output of the Spatial Agent in a loop until the spatial rule is accepted. Once the spatial rule is accepted, the spatial rule will apply to furniture placement.

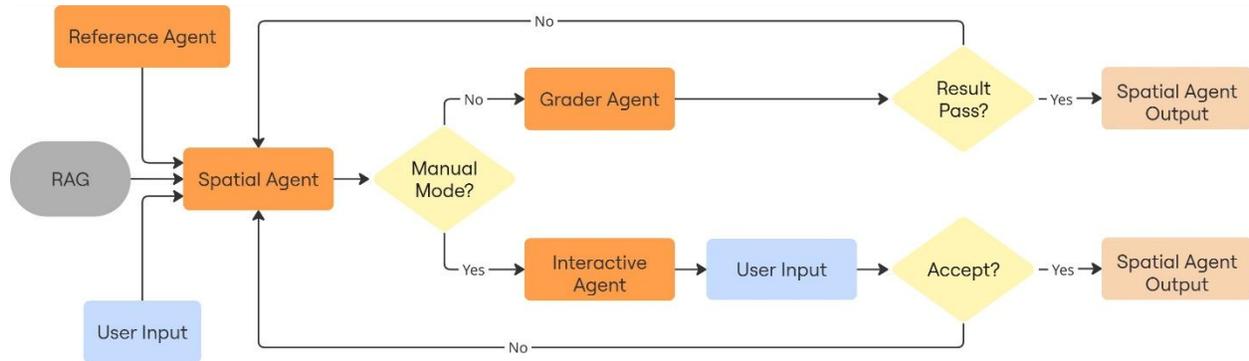

**Figure 3. Agent interaction diagram. User input, RAG context, and the Reference Agent seed the Spatial Agent, which generates initial spatial rules. Refinement follows either manual (user and Interactive Agent) or auto (Grader Agent loop) mode until final rules are accepted.**

### 3.3 Reference Agent

The Reference Agent converts reference images (Fig. 4) into structured text that seeds the Spatial Agent's decisions. A multimodal LLM analyzes each reference image (I_reference) with guiding prompt (T_guide), extracting information about furniture types and spatial relationships. Descriptions and images are stored in shared memory, providing context for subsequent rule generation. This foundational analysis provides critical contextual information guiding the decision-making process of the Spatial Agent.

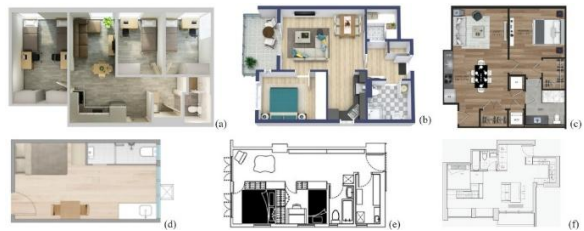

**Figure 4. (a) 63 m² university student apartment layout; (b) 47 m² one-bedroom, one-bathroom apartment layout; (c) 75 m² one-bedroom, one-bathroom apartment layout; (d) 15 m² transitional housing layout; (e) 32 m² apartment layout; (f) 33 m² studio apartment layout.**

## 3.4 Spatial Agent

The Spatial Agent selects furniture and formulates spatial rules, containing the core participatory element within the multi-agent framework. The Spatial Agent utilizes chain-of-thought pipelines to increase performance (Fig. 5). The inputs, including room type (R_type), room function (R_function), room size (R_size), room polygon (R_polygon), and reference text from the Reference Agent (T_reference), enter the first LLM, which outputs an object list (O_object = {t, a, f, q}), covering room type (t), object name (a), furniture factory (f), and quantity (q), all aligned with user preferences and room boundaries. A subset (O_objsub = {a, q}), combined with the original inputs, feeds two further LLM calls that generate placement rules. The final output comprises three parts: the previously mentioned (O_object); objects constraint (O_constraint = {a, c1, c2}), including object name (a), first constraint (c1), and second constraint (c2); and objects score terms (O_scoreterms = {a, d, acc, ang, f, v}), including object name (a), distance parameters (d), accessibility parameters (acc), angle alignment parameters (ang), focus score (f), and volume (v). Such strategy breaks tasks into steps for precise designs and converts text output to Python dictionaries for easier 3D scene generation.

Integration of RAG to the Spatial Agent is a key strength. Before generating outputs, the agent fetches domain guidelines from a local design-rule repository, injecting them as context. RAG therefore supplies expert design knowledge without massive task-specific training, boosting accuracy while remaining lightweight. By combining user input, chain-of-thought decomposition, and RAG, the Spatial Agent bridges technical design expertise and participatory goals, producing resilient layouts that flexibly adapt to diverse rooms and client requirements, within a data-efficient, easily computable format.

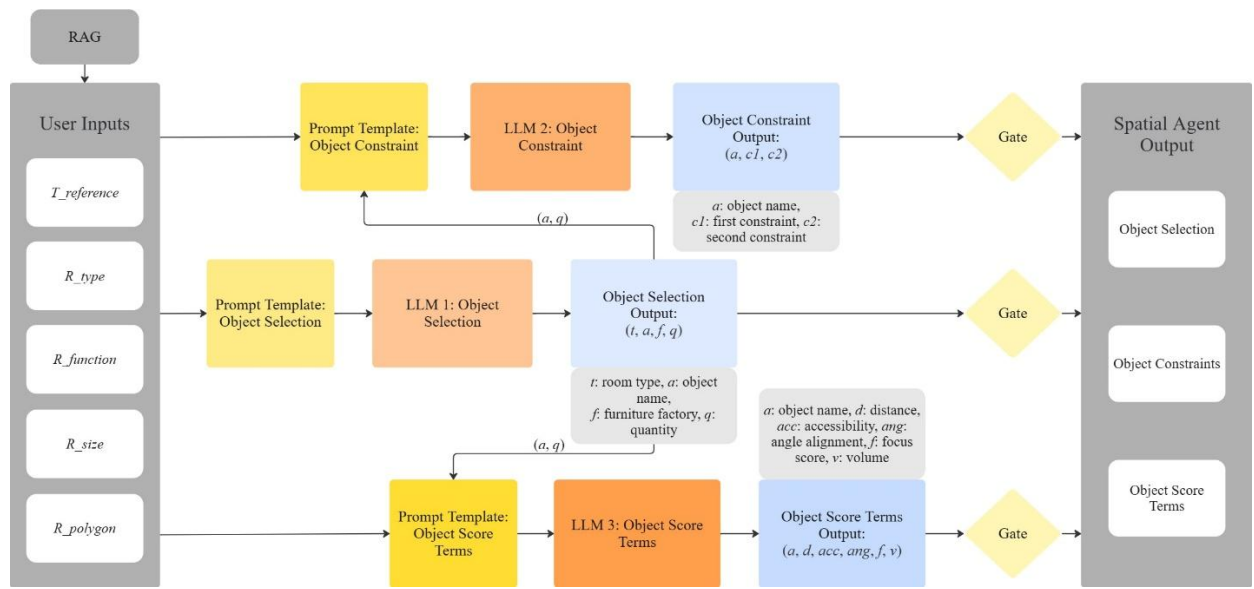

**Figure 5. The diagram of Spatial Agent. RAG and user inputs flow through three prompted LLM stages including object selection, object constraint, and object score terms. Each output from the stages passes a gate, the accepted results collectively form the agent's final lists of selected objects, associated spatial constraints and score terms.**

### 3.5 Interactive Agent

The Interactive Agent further emphasizes participatory design by transforming complex technical terms into plain language. The Interactive Agent takes the output from Spatial Agent as input (O_object, O_constraint, O_scoreterms), then converts each into user-friendly descriptions, and posts them to the front-end interface. There, users without design training can review, accept, or reject every proposal. Acceptance advances the workflow; rejection sends feedback to the Spatial Agent, which revises its output and resubmits it for retranslation until approval is reached. This real-time loop enables users without technical backgrounds to actively engage in the design process, ensuring their intent is captured. By lowering the jargon barrier, the Interactive Agent fosters meaningful dialogue and an inclusive, collaborative workflow.

### 3.6 Grader Agent

The Grader Agent delivers automated quality control. Leveraging the multimodal LLaVa-1.5-7B-hf, it receives room type (R_type) and the output of the Spatial Agent (O_object, O_constraint, O_scoreterms), compares these rules with reference images, and returns a score between 0 and 100. LLaVa's fine-grained vision reasoning—state of the art across 11 vision-language benchmarks—suits judging object selection and spatial relationships (Liu et al. 2024). Its lightweight, data-efficient, open-source model removes token costs in the iterative Spatial Agent to Grader Agent loop, whereas alternatives such as GPT-4V would be prohibitively expensive.

### 3.7 Prompt Guideline

Prompt engineering steers each LLM toward precise spatial-design output (White et al. 2023). Adapting the Template, Persona, and Fact-Checklist patterns, we built three structured prompts for the Spatial Agent (object selection, constraints, and score terms) and three parallel prompts for the Interactive Agent. All six templates rely on explicit placeholders, example-driven formatting, and rigid output restrictions, significantly reducing ambiguity and parse errors.

The prompt templates emphasize explicitness, tightly bounded options, and exemplar pairs that show correct versus incorrect examples. This design keeps results consistent, computer-readable, and responsive to varied user demands (all prompt templates in project page).

### 3.8 Application of the Framework to Interior Design

To demonstrate practical application of our framework, we integrated our proposed framework into Infinigen Indoors (Raistrick et al. 2024), an open-source, 3D synthetic scene generator. In our approach, the LLM Multi-Agent framework replaces Infinigen's hard-coded constraint language by generating user-centric spatial rules, while the built-in simulated annealing (SA) algorithm refines those rules into optimized furniture layouts. This combination of LLM generated spatial rules and SA optimization solver eliminates the costly data-training step and remains adaptable to any new furniture catalog and third-party 3D assets, hence improving both resilience and stability. This case study provides detailed insights into the operational effectiveness of our multi-agent framework.

**Initial User Input.** The design process begins with users defining input parameters. In this example, the framework will generate indoor design for a 22 m$^2$ living room with the floor plan named "rooms_1B1B," a floor plan taken from a publicly accessible catalog of residential layouts (Alea Miami n.d.), serving as a realistic reference for our case study. A user requirement "living room with dining function" is specified in the textbox. The generated design file will be stored in the folder "floorplan_1b1b." Then the user presses the "start generate" button to begin the generation pipeline. Once started, input parameters are saved as a JSON file to ease the communication between frontend and our framework. These inputs serve as foundational requirements for design, ensuring the generated design aligns with the user's demand.

**Interaction Between Multi-Agents and User.** Next, the Spatial Agent proposes an object selection list, while the Interactive Agent immediately paraphrases it to natural language. Both the raw output from "Spatial Agent Messages" and the plain language message from "Interactive Agent Messages" appear in the "AI Agent Chatbox" tab. At first, the user rejects and gives the feedback "I don't want any side tables and armchair. Add 3 plants to make room more vivid." The Spatial Agent proposes a new object selection list based on user feedback, then reparaphrases, redisplays, and the user accepts the new proposed object selection list. The same interactive loop repeats twice more: object constraint and object score-terms (Fig. 6). If, instead, the user toggles the "auto" button, a Grader Agent will simulate the user and refine the output of Spatial Agent automatically.

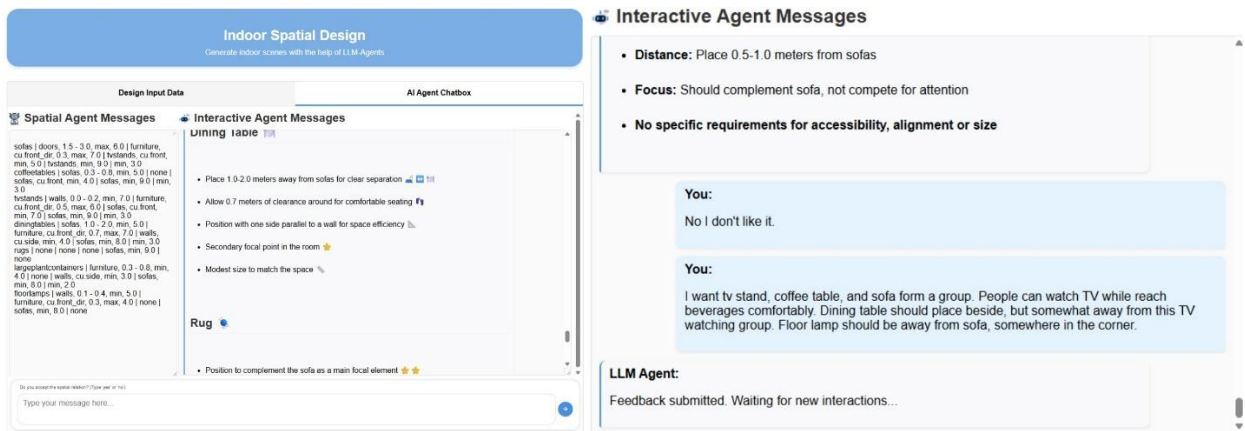

**Figure 6.** User interaction example: Object Score Terms stage in the framework

**Placement Optimization and Populate Objects.** A constraint graph is generated to define spatial relationships between objects (e.g., "nightstand placed next to the bed," "floor lamp 3 to 5 meters away from window"). The system then optimizes placements using simulated annealing with Metropolis-Hastings and the loss function and violation function is calculated based on the accepted object score terms and object constraints, respectively. The simulated annealing will run 80 iterations for every large object (parent objects, e.g., sofa), 60 iterations for every medium object (child objects, e.g., coffee table, which is related to sofa), and 30 iterations for every small object (decoration objects) to optimize object placement. Finally, 3D models of furniture are populated into the scene.

**3D scene and Reports.** The framework generates a realistic 3D render of the optimized indoor design blender file and supplementary reports (Fig. 7). Including metrics such as simulated annealing loss function graph and log files of the 3D layout generation process, offering transparency into the automated design process.

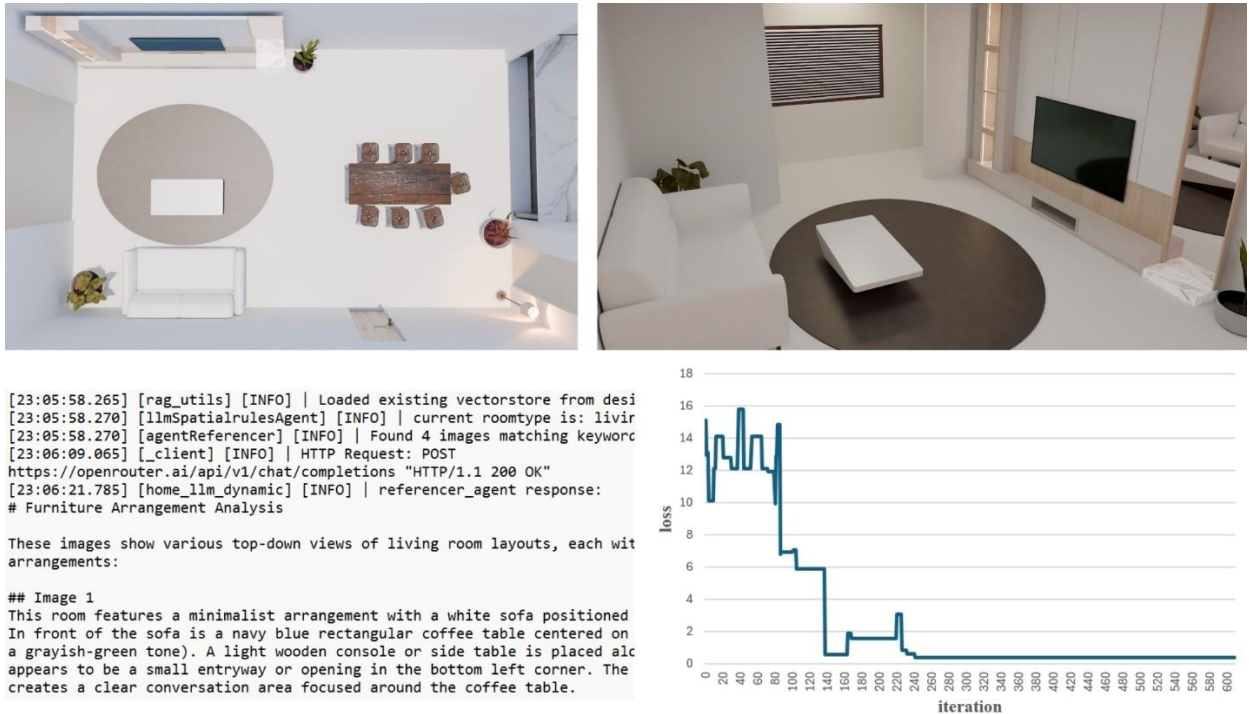

**Figure 7. Output files: 3D design file (top), log file of the generation process (bottom left), and loss graph of simulated annealing (bottom right).**

4.  Evaluation

To evaluate the effectiveness of participatory design in our framework, we conducted a comparative analysis of two 3D indoor scene generation methods: one with user interaction and one without. In the non-interactive method, the layout is generated solely based on the intrinsic spatial reasoning capabilities of the LLM, using identical room specifications and user requirements. This baseline illustrates the limitations of relying exclusively on the LLM's internal understanding of spatial relationships, which can sometimes diverge from human expectations. In contrast, the interactive method incorporates user feedback to iteratively refine spatial rules.

An independent LLM, Chat GPT-o3 (OpenAI n.d.), was employed to evaluate the resulting 3D scenes across four criteria: user intent alignment, aesthetic coherence, functionality, and circulation design. These criteria were selected as they collectively reflect core aspects of interior design quality that participatory input is expected to enhance. For each design, the LLM received a top-view render of the layout together with the four-criteria evaluation prompt. All evaluations were run in the OpenAI web interface with default interface settings. Using a single prompt that asked GPT-o3 to score the four criteria directly yielded nonsensical evaluation. To calibrate, we compiled a rubric of "good vs. bad" layout cues for each criterion by synthesizing standard interior-design guidance (e.g., clearances, scale/proportion, circulation paths), then integrated this rubric into the four-criteria evaluation prompt (Table 1). The use of an LLM for evaluation offers several advantages identified in recent work, including efficiency, promptable customization, and explainability, making it suitable for detailed comparison even in small-scale evaluation (Zhang et al. 2025). A user survey further contrasted satisfaction levels between our framework and existing tools, showing the enhancement of participatory design in our framework.

**Table 1: Example evaluation metric for the four grading criteria, adapted from the compiled evaluation rubric**

| CRITERIA | Good indicators | Bad indicators |
|---|---|---|
| User-intent alignment | Zoning, seating, storage, and lighting clearly support the stated activities and users (e.g., reading + movie-watching for a couple); seating oriented to the focal task; capacity and ergonomics match the brief. | Elements contradict the brief; insufficient seating; no task lighting or surfaces for reading; viewing axis blocked or mis-aligned. |
| Aesthetic coherence | Cohesive palette/materials; consistent style vocabulary; clear focal point; rhythm/repetition; pieces scaled in proportion to the room; intentional contrast. | Conflicting styles/colors; visual clutter; no focal point; disproportionate pieces that dominate or look "lost"; arbitrary accents. |
| Functionality | Furniture scale suits room; surfaces reachable; doors/windows unobstructed; plausible task/general lighting; storage accessible; placements enable intended use. | Oversized/undersized pieces; unreachable tables; blocked openings/windows; hazardous or awkward placements; inadequate lighting for stated tasks. |
| Circulation | Continuous paths between entry, seating, storage, and focal points; no pinch points; pass-through within groupings; seats usable without moving other items. | Routes blocked; tight pinch points; detours or dead ends created by clusters; chairs cannot slide out; circulation crosses hazards. |

### 4.1 Compact Living Room

A comparison between two living-room designs appears in Fig. 8. Both start from identical inputs: a 14 m² public-housing floor plan and the requirement "single occupant who enjoys coffee and TV." The interactive pipeline averaged 7.50 against the non-interactive baseline's 6.25. Gains were largest in circulation (5 to 7), followed by functionality (6 to 7) and intent alignment (7 to 8). For the design without user interaction, the system centered a sofa opposite the TV stand, squeezed a coffee table against it, and added an armchair that pinched the diagonal entry path, which is adequate but cramped. For the design with interaction, the client removed the armchair, shifted the coffee table to give leg clearance, and added a side table, shelf, and two plants. These edits opened an L-shaped circulation corridor from entry to seating, balanced beverage placement with screen sightlines, and enlivened the room without extra seating. The result meets ergonomic norms and the "coffee while watching TV" intent more closely, confirming the ethos of participatory adjustment.

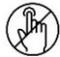

**Figure 8. Independent LLM evaluation of a compact living room**

### 4.2 Spacious Living Room

We applied both methods to a 23 m² apartment living room (Fig. 9), with requirements "a couple who enjoy books and movies." The result with interaction averaged 7.75 versus 6.50 for the one without interaction, confirming that participation boosts quality. User-intent alignment showed the biggest gain (8 to 6), while functionality rose 6 to 7 and both aesthetic coherence and circulation increased from 7 to 8. In the design without interaction, a sofa, armchair, and coffee table were bunched together, blocking passage. A TV shelf sat in a corner with no seating. The interactive design repositioned pieces to open an L-shaped path from entry to rooms, kept clear spacing around seats, and split the space into two zones: a movie area with sofa, coffee table, and TV shelf, and a reading corner with armchairs, side table, bookshelf, and floor lamp. Such arrangement delivers simultaneous movie watching and reading without crossover, turning a serviceable layout into a user-centric environment.

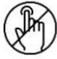

Figure 9. Independent LLM evaluation of a spacious living room

### 4.3 Micro Bedroom

We extended the comparison to an 8 m² public-housing bedroom with user requirements "for a man who likes to read." (Fig. 10). The result with interaction averaged 6.75 versus 5.50 for the one without; user-intent alignment increased from 6 to 7, aesthetic coherence increased from 7 to 8, functionality rose from 5 to 6, and circulation improved from 4 to 6. Even in a tight, rectangular room, participation raised quality across all metrics. In the design without interaction, a bookshelf with two lamps sat opposite an armchair. A bed lay mid-room, and a nightstand blocked the doorway, which technically satisfied the reading requirement but left cramped walkways. For the design with interaction, the armchair moved beside the bookshelf to form a compact reading space, the bed shifted against the wall, and the nightstand moved bedside, clearing an unobstructed path from door to window. These adjustments preserved every functional element while enhancing flow and ergonomics, showing that interactive design can refine even the most space-constrained layouts into comfortable, purpose-built environments.

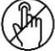

Figure 10. Independent LLM evaluation of a compact bedroom

### 4.4 User Survey

We evaluated the participatory strength of our LLM-based, multimodal, multi-agent framework through an eight-question questionnaire (survey questions in project page) delivered through the web UI prototype. Questions 1–7 addressed five key aspects of participatory design: sense of co-creation, transparency and explainability, expressiveness and ease-of-use, collaboration friendliness, and responsiveness to feedback; Question 8 asked respondents which tool they would choose for a real project, selecting between our framework and traditional design software (e.g. SketchUp and RoomSketcher).

Among the 53 valid responses (Fig. 11), 77% of participants reported being "satisfied" or "very satisfied" with the participatory qualities of the framework. About 89% agreed that it helped them express preferences without technical jargon. For tool preference, 51% favored our framework, 42% indicated no strong preference or wished to combine it with conventional software, and only 8% selected traditional tools.

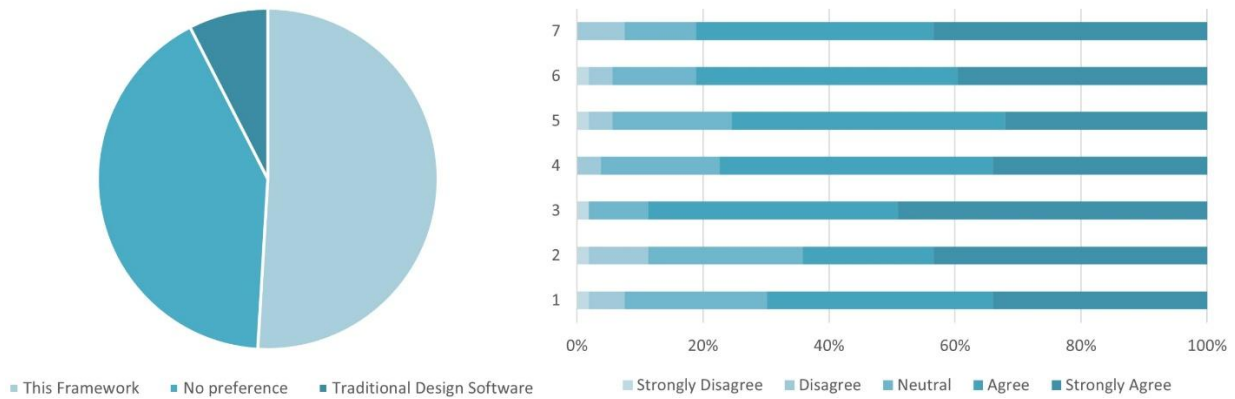

**Figure 11. Questionnaire result of 53 participants, about overall tool preference (left) and rating of the framework about participatory design elements (right)**

Both participants with and without professional design knowledge show support for our framework (Fig. 12). The data shows a balanced distribution across occupations. Among those with formal design expertise (e.g., designers and architects; n = 7), 71% indicated they would choose our framework for an actual project, with the remainder largely neutral, preferring both our framework and traditional design software. At the same time, nondesigners, including healthcare professionals (n = 7) and engineers (n = 5), also expressed clear favor for our framework, with 71% of healthcare professionals and 60% of engineering and technology professionals preferring our framework, and only a small minority choosing traditional design software. Interestingly, many participants who gave high ratings nevertheless reported "no preference" between design tools (42%). This suggests that while people appreciate our framework's participatory features, some may default to familiar software out of habit and be cautious about adapting to new ways of designing. Furthermore, some designers express that while AI-driven design frameworks could effectively improve efficiency, they still wish to lower the level of automation. They would like the option to let AI handle only repetitive tasks, as many designers still enjoy being more hands-on.

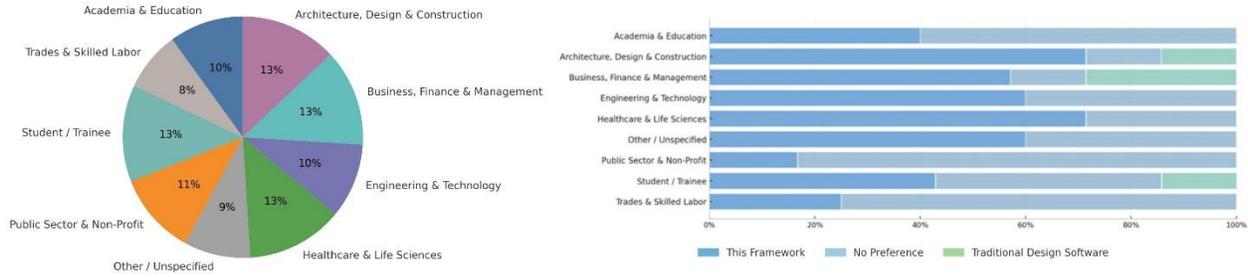

**Figure 12. Occupation distribution of 53 participants (left) and tool preference distribution by occupation group (right)**

Overall, most of the participants hold a positive view of our framework, with strong cross-disciplinary endorsement from designers, healthcare professionals, and engineers. Yet 42% still indicated no strong preference, suggesting many default to familiar tools. Several designers also asked for finer-grained automation control to preserve hands-on involvement. However, with only 53 participants, these results should be validated in larger, more diverse investigations.

## 5. Conclusion

This work bridges the gap between 3D scene synthesis and participatory design by integrating LLM based multi-agent collaboration. Our framework addresses three main challenges: converting nondesigner intent into spatial rules, enhancing creative efficiency for designers, and demonstrating practical value through real-world case studies. Evaluation revealed its ability to generate quality interior design layouts while reducing reliance on costly training datasets. By enabling stakeholders to iteratively refine designs through natural language, the framework makes creativity in architectural design more accessible without compromising technical accuracy, contributing to both the theory of resilient AI agent systems and practical inclusive design workflows.

However, limitations remain. User feedback highlighted needs for enhanced visual interaction (e.g., drag-and-drop interaction, real-time 3D previews) and multilingual support to broaden accessibility. Some design professionals suggested that the text-heavy interface might discourage users who are more familiar with visual graphical tools. Additionally, depending on simulated annealing optimization restricts adaptability to unconventional spatial requirements. Hence, future work aims to prioritize multimodal interactions, including sketches and mood boards, and add multilingual capabilities to enhance inclusivity; extend the framework to broader architectural scenarios such as urban planning and architecture design; and improve the placement algorithm to create more efficient design workflows.

In conclusion, this research underscores the potential of LLM-based agent systems in architecture. Our framework not only improves design efficiency but also redefines creativity as a collective endeavor. As AI evolves, such a participatory system could democratize high-quality design, ensuring architectural innovation serves both experts and nonexperts.

**Acknowledgements:**

This work was supported by the Hong Kong Center for Construction Robotics, an InnoHK center funded by the Innovation and Technology Commission of the HKSAR Government.

To improve clarity and grammatical accuracy, we used ChatGPT (OpenAI, version 4.5, 2025) solely for copy-editing purposes.

**Notes:**

Prompt templates and the survey questions used in this study are available on the project website: https://rsigktyper.github.io/AICodesign/

**Image Credits:**

Figure 4: © (a) University of Miami Housing & Residential Life, (b) RoomSketcher, (c) Bezel at Miami Worldcenter, (d) Hong Kong Housing Bureau, (e) Meunier (2011), (f) Vetterlein (2020).

All other drawings and images by the authors.

# Intelligent Co-Design: An Interactive LLM Framework for Interior Spatial Design via Multi-Modal Agents: Appendix

## Appendix A: Prompt Templates

Object Selection Prompt:

You are an interior designer selecting furniture for a {room_type}.
Strictly follow the below instructions!

## AVAILABLE FACTORIES
seating.BedFactory, seating.SofaFactory, seating.ArmChairFactory,
seating.ChairFactory, seating.OfficeChairFactory, seating.BarChairFactory
tables.CoffeeTableFactory, tables.SideTableFactory, tables.TableDiningFactory
lamp.FloorLampFactory shelves.TVStandFactory, shelves.SimpleBookcaseFactory, shelves.CellShelfFactory,
shelves.LargeShelfFactory, shelves.KitchenCabinetFactory, shelves.SingleCabinetFactory
appliances.DishwasherFactory, appliances.OvenFactory, appliances.BeverageFridgeFactory
bathroom.StandingSinkFactory, bathroom.ToiletFactory, bathroom.BathtubFactory
elements.RugFactory

## GUIDELINES
1. Choose furniture suited for a {room_type} of room size: {room_size} square meters and polygon {room_polygon}.
2. The {room_type} has this specification: {room_spec}.
3. Object names must be **lowercase plural without spaces** (e.g., coffeetables, floorlamps)
4. Include quantities for multiple identical items.
5. Follow these constraints:
- Maximum: 1 floor lamp per room.
- Scale furniture to fit {room_size} square meters.

## SUGGESTION
1. If room size is small (e.g. 10-20 square meters), choose smaller furniture.
2. Choose less furniture if room size is small.

## TEMPLATE
room_type | selected_objects | furniture_factory | quantity

## OUTPUT EXAMPLES
livingroom | sofas | seating.SofaFactory | 1
livingroom | coffeetables | tables.CoffeeTableFactory | 1

Provide only the formatted output with no additional text.

Figure 13. Spatial Agent Prompt Template (Object Selection Prompt)

> Object Constraints Prompt:
>
> You are designing a {room_type} (polygon: {room_polygon}) with these 'objects': {selected_objects}
> Strictly follow the below instructions!
>
> ## CONSTRAINTS
> 1. **Global Positioning** (MUST ONLY related to 'rooms'):
>    none (this mean on floor already, freestanding), against_wall, corner_against_wall
>    flush_wall, spaced_wall, side_against_wall, back_near_wall, side_near_wall,
>    CORRECT Example: "rooms, against_wall", "rooms, none", "rooms, back_near_wall".
>    INCORRECT Example: "sofas, side_near_wall", "tvstands, back_near_wall".
>
> 2. **Object-to-Object Positioning** (MUST ONLY related to 'objects'):
>    none, front_against (front against three sides), front_to_front (front against front), leftright_leftright (left and right against two sides), side_by_side (side against side), back_to_back (back against back)
>    CORRECT Example: "sofas, front_to_front", "beds, side_by_side".
>    INCORRECT Example: "sofas, side_near_wall", "tvstands, back_near_wall".
>    NOTE: 'objects' refers to anything in {selected_objects}.
>
> ## GUIDELINES
> 1. Use exactly the object names in {selected_objects}.
> 2. The {room_type} has this specification: {room_spec}.
> 2. Each object type only have one type of positioning, either Global Positioning or Object-to-Object Positioning.
> 3. Each positioning includes 1-2 constraints per object type related to the related parent object.
> 4. Start with anchor objects and place larger objects first.
> 5. Maximize open space (place objects against walls).
> 6. Rugs should always be: `rugs | rooms, none`.
>
> ## TEMPLATE
> selected_objects | constraint_1 | constraint_2 | ...
>
> ## OUTPUT EXAMPLES
> beds | rooms, against_wall
> beds | rooms, corner_against_wall
> tvstands | rooms, against_wall
> coffeetables | sofas, front_to_front
> nightstands | beds, leftright_leftright
> ### DON'T have different related parents like this:
> tvstands | rooms, against_wall | sofas, front_against
>
> Provide only the formatted output with no additional text.

**Figure 14. Spatial Agent Prompt Template (Object Constraints Prompt)**

Object Score Terms Prompt:

You are designing a {room_type} (polygon: {room_polygon}) of room size: {room_size} square meters with objects: {selected_objects}

Utilize more on 'Focus Score', less on 'Volume'

## AVAILABLE SCORE TERMS:
1. **Distance**: distance(related_object, distance_range, min_or_max, weight)
   - Example: "doors, 1.5 - 3.0, max, 8.0" (maximize distance to doors, weight=8.0)
   - Example: "sofa, 0.5 - 1.0, min, 1.0" (minimize distance to sofa, weight=1.0)
   - Note: distance_range can be "none" to indicate no specific range
2. **Accessibility**: accessibility(related_object, direction, distance, min_or_max, weight)
   - Direction: cu.front_dir, cu.back_dir, cu.down_dir
   - CORRECT Example: "furniture, cu.front_dir, 0.3, max, 8.0"
   - INCORRECT Example: "furniture, cu.side_dir, 0.5, max, 4.0"
   Note: MUST choose from items listed in 'Direction'
3. **Angle Alignment**: angle_alignment(related_object, orientation, min_or_max, weight)
   - Orientation: cu.front, cu.side, cu.back, cu.top, cu.leftright, cu.bottom
   - Example: "tvstand, cu.front, min, 6.3"
4. **Focus Score**: focus_score(related_object, min_or_max, weight)
   - Example: "dinningtable, min, 7.0"
5. **Volume**: volume(min_or_max, weight)
   - Example: "min, 5.0"

## GUIDELINES
1. Use exactly the object names in {selected_objects}. WITHOUT OTHER OBJECTS.
2. The {room_type} has this specification: {room_spec}.
3. **Output Sequence**: The output must strictly follow the sequence: `selected_objects | distance | accessibility | angle_alignment | focus_score | volume`.
4. Parent object should not have score terms related to child object, only same level of parent objects.
5. Assign 1-3 score terms per object type.
6. Weights range from 0.0 to 10.0.
7. Additional related objects: doors, windows, furniture, opens.
8. Example rules:
   - Place seats/beds far away from doors with high weight.
   - Chairs should align to face tables.
   - Keep furniture from blocking doors/windows/other furniture.
   - Ensure front accessibility for storage/seating.
   - Use "none" for rugs and unused terms
9. Should always make objects compact.

## TEMPLATE (MUST be in this order)
selected_objects | distance | accessibility | angle_alignment | focus_score | volume

## CORRECT OUTPUT EXAMPLES
sofas | doors, 1.5 - 3.0, max, 5.0 | furniture, cu.front_dir, 0.1, max, 6.0 | none | none | min, 8.0
coffeetables | none | none | sofas, cu.front, min, 5.0 | sofas, min, 6 | none
nightstands | walls, 0.0 - 0.3, min, 8.0 | none | beds, cu.front, min, 6.0 | none | none
rugs | none | none | none | none | none

## INCORRECT OUTPUT EXAMPLES
floorlamps | furniture, cu.front_dir, 0.5, max, 2.0 | none | none | none | none
sofas | doors, 1.5 - 3.0, max, 5.0; windows, 1.0 - 2.0, max, 4.0 | furniture, cu.front_dir, 1.0, max, 6.0 | none | none | min, 8.0
beds | doors, 1.5 - 3.0, max, 5.0; windows, none, max, 2.0 | none | none | none | min, 7.0
simpleshelves | furniture, cu.front_dir, 0.3, max, 6.0 | none | none | none | none
armchairs | doors, 0.5 - 1.5, max, 7.0 | none | sidetables, cu.side, min, 5.0 | none | none
coffeetables | sofas, 0.01 - 0.2, min, 4.0 | none | sofas, cu.front, min, 5.0 | min, 6 | none

Provide only the formatted output with no additional text.

**Figure 15. Spatial Agent Prompt Template (Object Score Terms Prompt)**

> Object Selection Prompt:
>
> You are a professional interior design translator who makes technical furniture specifications easy to understand for clients.
>
> Convert this technical furniture list into simple, easy-to-read sentences. You should give some icons for better visualization.
>
> The technical list has this format:
> room_type | selected_objects | furniture_factory | quantity
>
> For example:
> "livingroom | sofas | seating.SofaFactory | 2" means "For the living room, I've selected 2 sofas."
> "bedroom | beds | seating.BedFactory | 1" means "For the bedroom, I've selected 1 bed."
>
> Technical list:
> {raw_object_selection_text}
>
> Please convert each line to a simple sentence explaining what furniture was selected for which room and in what quantity. Ignore the factory names in your explanation.

**Figure 16. Interactive Agent Prompt Template (Object Selection Prompt)**

> Object Constraints Prompt:
>
> You are a professional interior design translator who makes technical furniture specifications easy to understand for clients.
>
> Convert these technical furniture placement constraints into simple, easy-to-understand explanations, keep the explaination short and brief. You should give some icons for better visualization.
>
> The constraints are in this format:
> selected_object | constraint_1 | constraint_2 | ...
>
> Where constraints can be:
> - Global Positioning (related to rooms): none, against_wall, flush_wall, spaced_wall, side_against_wall, back_near_wall, side_near_wall
> - Object-to-Object Positioning (related to other furniture): none, front_against, front_to_front, leftright_leftright, side_by_side, back_to_back
>
> For example:
> - "beds | rooms, against_wall | rooms, side_near_wall" means the bed should be positioned against a wall with one side also near a wall.
> - "coffeetables | sofas, front_to_front" means the coffee table should be placed with its front facing the front of the sofa.
>
> Input constraints:
> {raw_constraints_text}
>
> Please translate each line into a clear explanation of how each piece of furniture should be positioned in the room. Use everyday language that describes the practical placement, like:
> - "The bed will be placed against a wall with one side also near a wall for better space utilization."
> - "The coffee table will be centered in front of the sofa."
>
> Organize your response by furniture type and avoid technical terms like "front_against" or "side_near_wall" - instead, describe what this means visually in the room.

**Figure 17. Interactive Agent Prompt Template (Object Constraints Prompt)**

> Object Score Terms Prompt:
>
> You are a professional interior design translator who makes technical furniture specifications easy to understand for clients.
>
> Convert these technical furniture optimization scores into simple, easy-to-understand explanations, keep the explaination short and brief. You should give some icons for better visualization.
>
> The score terms are in this format:
> selected_objects | distance | accessibility | angle_alignment | focus_score | volume
>
> Where:
> - Distance: How far furniture should be from other objects
> - Accessibility: How much clear space is needed in front/behind furniture
> - Angle Alignment: How furniture should be oriented relative to other pieces
> - Focus Score: How much attention a piece should draw in the room
> - Volume: Whether to minimize/maximize the size of furniture
>
> Note: explain all numbers in meters
>
> For example:
> "sofas | doors, 1.5 - 3.0, max, 5.0 | furniture, cu.front_dir, 0.1, max, 6.0 | none | none | min, 8.0"
> means the sofa should be placed far from doors (1.5-3.0 meters), needs a small amount of clear space in front (0.1 meters), and should be compact in size.
>
> Input score terms:
> {raw_scoreterms_text}
>
> Please translate each line into a friendly explanation of how each piece of furniture will be optimally placed in the room. Explain:
> 1. Its relationship to other furniture and room elements (distance)
> 2. How much space is kept clear around it (accessibility)
> 3. Which way it faces or how it's oriented (angle alignment)
> 4. Whether it's a focal point in the room (focus score)
> 5. Size considerations (volume)
>
> Use everyday language a non-designer would understand. Organize by furniture type and explain why each consideration improves the room layout.

**Figure 18. Interactive Agent Prompt Template (Object Score Terms Prompt)**

## Appendix B: Evaluation Setup

> Evaluation Prompt Example:
>
> You are an interior design evaluator.
>
> Your will learn from the design criteria rubric, understand what is a good design and bad design, then you will grade the generated 3D scene.
>
> Design criteria rubric:
> {design_criteria_rubric}
>
> Please grade the following 3D indoor design base on the following criteria, give a score out of 10 for each criteria, and give a brief reason why you give this score:
> 1. User intent alignment
> 2. aesthetic coherence
> 3. functionality
> 4. circulation design
>
> Below is the room specification and user requirement:
> room type: {room_type}
> room spec: {room_spec}
> room size: {room_size}

**Figure 19. Evaluation Prompt Example**

## Appendix C: Survey Questionnaire

> For questions 1-7, please rate from 1 (Strongly Disagree) to 5 (Strongly Agree).
>
> Sense of Co-Creation
> 1. I felt we were co-creating the design rather than the framework designing for me.
>
> Transparency & Explainability
> 2. The framework clearly explained how each design choice was made.
>
> Expressiveness & Ease-of-Use
> 3. The framework helped me express my preferences without requiring technical jargon.
>
> Overall Satisfaction
> 4. Overall, I am satisfied with this framework for participatory design.
>
> System Comparison
> 5. This framework was easier to communicate compared to traditional design software.
> 6. This system felt more like a collaboration compared with traditional design software.
>
> Responsiveness to Feedback
> 7. The system adapted well to my feedback and design preferences.
>
> Preferred Tool
> 8. Overall, which one would you prefer to use in a real project, this framework or traditional design software (e.g. SketchUp and RoomSketcher)?

**Figure 20. Eight-question Survey Questionnaire**